\documentclass[10pt,twocolumn,letterpaper]{article}
\usepackage[pagenumbers]{cvpr}
\usepackage{tabularx,placeins,pgfplots}
\pgfplotsset{compat=1.18}
\definecolor{ink}{HTML}{19364B}
\definecolor{scrteal}{HTML}{007E83}
\definecolor{warm}{HTML}{B4661F}
\definecolor{pale}{HTML}{D9E0E5}
\definecolor{cvprblue}{rgb}{0.21,0.49,0.74}
\usepackage[pagebackref,breaklinks,colorlinks,allcolors=cvprblue]{hyperref}
\def\confName{CVPR}
\def\confYear{2027}

\newcommand{\sourcefigure}[2]{%
  \IfFileExists{#1.png}{\includegraphics[width=.90\textwidth]{#1.png}}{%
    \IfFileExists{#1(1).png}
      {\includegraphics[width=.90\textwidth]{#1(1).png}}{#2}}}

\title{When to Refine: Selective Search for Frozen Visual World Models}
\author{Jingni Huang \\ University of Oxford \\ \nolinkurl{jingni.huang@kellogg.ox.ac.uk, jingnih@gmail.com}}

\begin{document}
\maketitle

\begin{abstract}
Planning with a visual world model requires deciding both which actions
to execute and how long to search for them. We introduce Selective CEM
Refinement (SCR), a learned rule for deciding whether to continue
cross-entropy method (CEM) search with a frozen LeWorldModel. After
five iterations, the gate measures how far the mean first action chunk
has moved, how much the mean terminal cost of elite plans has fallen,
and how widely their first chunks vary. It returns the current plan
for execution or continues the same search for ten more iterations.
The gate requires no additional world-model rollout.

On 16 PushT scenes evaluated with two planning seeds, SCR succeeded
in 25 of 32 trials, compared with 28 of 32 for fixed 15-iteration
search (Fixed-15). All 25 SCR successes occurred in trials that
Fixed-15 also solved, yielding $89.29\%$ paired success retention.
Across complete episodes, including replanning, SCR evaluated
$23.81\%$ fewer candidate plans and reduced mean controller time
per eight-environment batch by $17.22\%$. Its observed success rate
was 5.36 percentage points above an analytic mixture of the two
fixed-depth controllers at the same expected candidate workload,
although the difference changed sign across planning seeds.
Compared with gating only at the initial solve under the same
random-draw padding protocol, gating again at replanning produced
nine additional early exits. This reduced total candidate evaluations
by $15.79\%$, while successes fell from 27 to 25.
\end{abstract}

\section{Introduction}
\label{sec:intro}

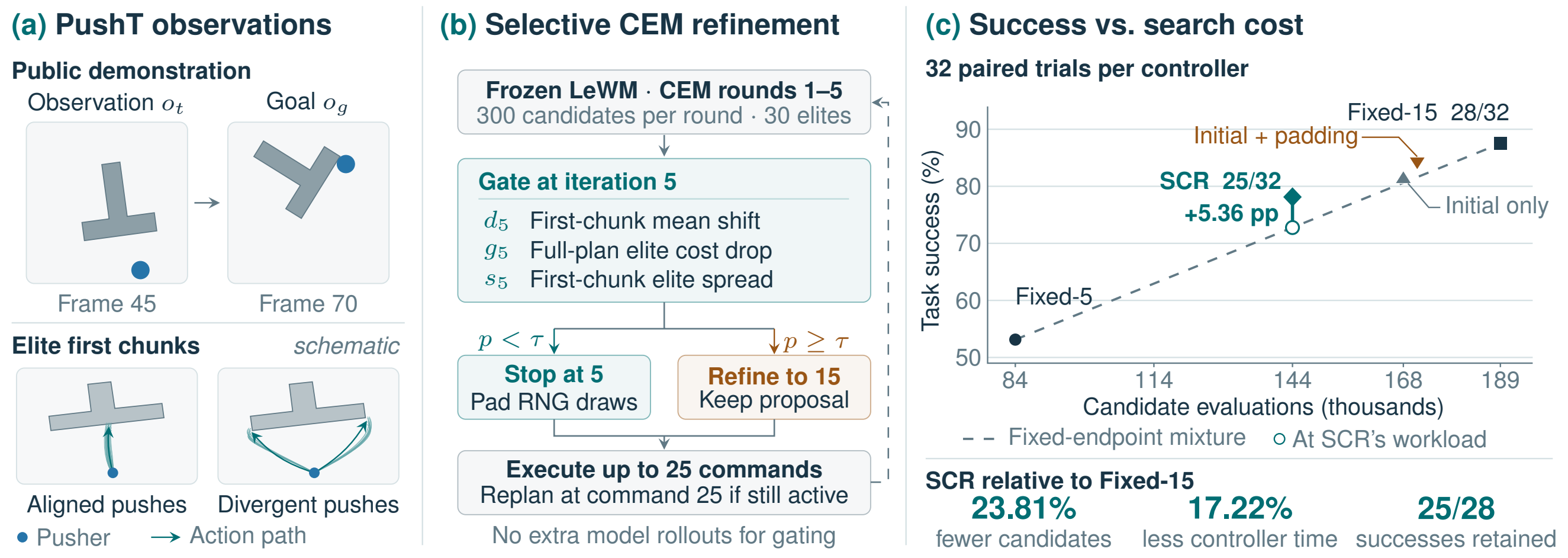
\begin{figure*}[t]
\centering
\resizebox{\textwidth}{!}{%
\begin{tikzpicture}[
  x=1mm,y=-1mm,>=stealth,
  font=\sffamily\fontsize{6.8}{7.5}\selectfont,
  every node/.style={inner sep=0pt,outer sep=0pt},
  title/.style={anchor=west,font=\sffamily\bfseries\fontsize{8.4}{9}\selectfont},
  lefttxt/.style={anchor=west},
  card/.style={rounded corners=.8mm,line width=.32pt},
  flow/.style={draw=bluegray,line width=.45pt}
]
\definecolor{deepink}{HTML}{183445}
\definecolor{sea}{HTML}{006E73}
\definecolor{ochre}{HTML}{995515}
\definecolor{bluegray}{HTML}{617985}
\definecolor{rulegray}{HTML}{D6E1E4}
\definecolor{pusher}{HTML}{2674A9}
\color{deepink}

\path[use as bounding box] (0,0) rectangle (159,55.6);
\draw[rulegray] (42.8,.2)--(42.8,55) (92,.2)--(92,55);
\node[title] at (1,2.3)    {\textcolor{sea}{(a)} PushT observations};
\node[title] at (44.6,2.3) {\textcolor{sea}{(b)} Selective CEM refinement};
\node[title] at (94,2.3)   {\textcolor{sea}{(c)} Success vs.\ search cost};

\node[lefttxt] at (1,6.7) {\textbf{Public demonstration}};
\foreach \x/\lab/\frameid in
  {10.7/{Observation $o_t$}/45,31.2/{Goal $o_g$}/70}{
  \node at (\x,10.1) {\lab};
  \draw[card,draw=rulegray,fill=bluegray!6]
    (\x-8.2,12) rectangle ++(16.4,16.4);
  \node[bluegray] at (\x,30.4) {Frame \frameid};
}
\def\drawpose#1#2#3{%
  \begin{scope}[shift={(#1,12)}]
    \filldraw[fill=bluegray!72,draw=bluegray,line width=.3pt]
      plot coordinates {#2}--cycle;
    \fill[pusher] #3 circle[radius=.93mm];
  \end{scope}}
\drawpose{2.5}{
  (13.27,11.05)(5.91,12.04)(5.66,10.20)(8.42,9.83)
  (7.68,4.31)(9.52,4.06)(10.26,9.58)(13.02,9.21)
}{(11.61,15.04)}
\drawpose{23}{
  (11.41,3.48)(7.48,9.78)(5.90,8.80)(7.38,6.43)
  (2.65,3.49)(3.63,1.91)(8.36,4.86)(9.83,2.49)
}{(12.01,4.26)}
\draw[flow,->] (19.6,20.2)--(22.2,20.2);

\draw[rulegray] (1,32.4)--(41,32.4);
\node[lefttxt] at (1,34.7) {\textbf{Elite first chunks}};
\node[anchor=east,bluegray] at (40.5,34.7) {\textit{schematic}};
\foreach \x/\kind in {1.5/0,22/1}{
  \begin{scope}[shift={(\x,37)}]
    \draw[card,draw=rulegray,fill=bluegray!5]
      (0,0) rectangle (18.4,11.8);
    \filldraw[fill=bluegray!45,draw=bluegray,line width=.3pt]
      (3.3,4.9)--(7.6,4.4)--(7.25,1.65)--(9.6,1.4)
      --(10,4.1)--(14.8,3.6)--(15,5.05)--(3.5,6.35)--cycle;
    \foreach \i in {1,...,30}{
      \pgfmathsetmacro{\d}{(\i-15.5)/30}
      \ifnum\kind=0
        \def\e{9.4}\def\h{5.95}\def\bend{-.35}
      \else
        \ifnum\i<16
          \def\e{3.4}\def\h{6.5}\def\bend{-1.3}
        \else
          \def\e{15.15}\def\h{5.35}\def\bend{1.3}
        \fi
      \fi
      \draw[sea!60,opacity=.70,line width=.15pt] (9.8,10.65)
        .. controls ({9.8+\bend+1.4*\d},9.7)
        and ({\e+\bend+.35*sin(69*\i)},7.3)
        .. ({\e+.6*\d},{\h+.42*cos(47*\i)});
    }
    \ifnum\kind=0
      \draw[sea,->] (9.8,10.65) to[bend left=9] (9.4,5.95);
    \else
      \draw[sea,->] (9.8,10.65) to[bend left=15] (3.4,6.5);
      \draw[sea,->] (9.8,10.65) to[bend right=13] (15.15,5.35);
    \fi
    \fill[pusher] (9.8,10.65) circle[radius=.54mm];
  \end{scope}
}
\node at (10.7,51.1) {Aligned pushes};
\node at (31.2,51.1) {Divergent pushes};
\fill[pusher] (2.1,54.2) circle[radius=.53mm];
\node[lefttxt,bluegray] at (3.5,54.2) {Pusher};
\draw[sea,->] (15.2,54.2)--(18.2,54.2);
\node[lefttxt,bluegray] at (19.1,54.2) {Action path};

\def\drawcard#1#2#3#4#5{%
  \draw[card,draw=#5!49,fill=#5!6] (#1,#2) rectangle (#3,#4);}
\drawcard{46.4}{6.7}{88.4}{13.2}{bluegray}
\node at (67.4,8.7) {\textbf{Frozen LeWM $\cdot$ CEM rounds 1--5}};
\node[bluegray] at (67.4,11.5) {300 candidates per round $\cdot$ 30 elites};
\draw[flow,->] (67.4,13.3)--(67.4,15.5);

\drawcard{46.4}{15.7}{88.4}{30.3}{sea}
\node[lefttxt,sea] at (48.5,18.0) {\textbf{Gate at iteration 5}};
\draw[sea!25] (48.5,19.8)--(86.2,19.8);
\foreach \y/\sym/\descr in {
  22/d_5/First-chunk mean shift,
  25.2/g_5/Full-plan elite cost drop,
  28.2/s_5/First-chunk elite spread}{
  \node[sea] at (50.4,\y) {$\sym$};
  \node[lefttxt] at (53.7,\y) {\descr};
}
\draw[flow] (67.4,30.3)--(67.4,33) (56.2,33)--(78.6,33);
\draw[sea,->] (56.2,33)--(56.2,35.6);
\draw[ochre,->] (78.6,33)--(78.6,35.6);
\node[anchor=east,sea] at (55.3,34.3) {$p<\tau$};
\node[lefttxt,ochre] at (79.5,34.3) {$p\geq\tau$};

\drawcard{46.4}{35.7}{66.0}{42.2}{sea}
\drawcard{68.8}{35.7}{88.4}{42.2}{ochre}
\node[sea] at (56.2,37.8) {\textbf{Stop at 5}};
\node at (56.2,40.4) {Pad RNG draws};
\node[ochre] at (78.6,37.8) {\textbf{Refine to 15}};
\node at (78.6,40.4) {Keep proposal};
\draw[flow] (56.2,42.2)--(56.2,43.9)--(78.6,43.9)--(78.6,42.2);
\draw[flow,->] (67.4,43.9)--(67.4,45.3);

\drawcard{46.4}{45.4}{88.4}{51.9}{bluegray}
\node at (67.4,47.5) {\textbf{Execute up to 25 commands}};
\node at (67.4,50.1) {Replan at command 25 if still active};
\draw[flow,dashed,->] (88.4,48.6)--(90.2,48.6)--(90.2,10)--(88.5,10);
\node[bluegray] at (67.4,54.2) {No extra model rollouts for gating};

\node[lefttxt] at (94,6.7) {\textbf{32 paired trials per controller}};
\begin{scope}[shift={(100.3,36.5)},x=.47mm,y=.58mm]
  \begin{scope}[shift={(-78,-49)}]
    \foreach \y in {50,60,70,80,90}{
      \draw[rulegray] (78,\y)--(196,\y);
      \node[anchor=east,bluegray] at (76.7,\y) {\y};
    }
    \draw[bluegray] (78,94)--(78,49)--(196,49);
    \foreach \x in {84,114,144,168,189}{
      \draw[bluegray] (\x,49)--(\x,48.3);
      \node[anchor=north,bluegray] at (\x,47.6) {\x};
    }

    \draw[bluegray,dashed,line width=.65pt] (84,53.125)--(189,87.5);
    \draw[sea,line width=.95pt] (144,72.767857)--(144,78.125);
    \draw[sea,fill=white,line width=.55pt]
      (144,72.767857) circle[radius=.66mm];

    \fill[deepink] (84,53.125) circle[radius=.62mm];
    \node[fill=deepink,minimum size=1.28mm] at (189,87.5) {};
    \filldraw[fill=bluegray,draw=white,line width=.23pt]
      (168,81.25) ++(0,1.5)--++(-1.75,-2.25)--++(3.5,0)--cycle;
    \filldraw[fill=ochre,draw=white,line width=.23pt]
      (171,84.375) ++(0,-1.5)--++(-1.75,2.25)--++(3.5,0)--cycle;
    \fill[sea] (144,78.125) ++(-2,0)
      --++(2,1.7)--++(2,-1.7)--++(-2,-1.7)--cycle;

    \node[lefttxt] at (84,60.7) {Fixed-5};
    \node[anchor=east] at (191,93.1) {Fixed-15\ \ 28/32};
    \draw[ochre] (171,85.2)--(167,88.6)--(159,88.6);
    \node[anchor=east,ochre] at (158.3,88.6) {Initial + padding};
    \draw[bluegray] (169,80.5)--(173.5,76.3)--(176,76.3);
    \node[lefttxt,bluegray] at (177,76.3) {Initial only};
    \node[anchor=east,sea] at (141.6,81) {\textbf{SCR\ \ 25/32}};
    \node[anchor=east,sea] at (141.1,75) {\textbf{+5.36 pp}};
  \end{scope}
\end{scope}
\node[rotate=90] at (94.7,24) {Task success (\%)};
\node at (128.4,41.1) {Candidate evaluations (thousands)};
\draw[bluegray,dashed,line width=.65pt] (97.8,44.2)--(101.5,44.2);
\node[lefttxt,bluegray] at (102.4,44.2) {Fixed-endpoint mixture};
\draw[sea,fill=white] (130,44.2) circle[radius=.53mm];
\node[lefttxt,bluegray] at (131.3,44.2) {At SCR's workload};

\draw[rulegray] (94,46.2)--(157.8,46.2);
\node[lefttxt] at (94,48.4) {\textbf{SCR relative to Fixed-15}};
\foreach \x/\val/\descr in {
  104/23.81\%/fewer candidates,
  126/17.22\%/less controller time,
  148/25\slash28/successes retained}{
  \node[sea,font=\sffamily\bfseries\fontsize{9}{10}\selectfont]
    at (\x,51.2) {\val};
  \node[bluegray] at (\x,54.3) {\descr};
}
\end{tikzpicture}%
}
\caption{Selective CEM refinement on PushT.
(a) Object and pusher poses are redrawn from frames 45 and 70 of
episode 0 in the public \texttt{lerobot/pusht\_keypoints}
demonstration data for PushT~\cite{pusht}. The first-chunk paths
are schematic.
(b) SCR decides once per solve whether to run ten further CEM
iterations.
(c) Filled markers show 32 paired trials per controller. The open
circle is the fixed-endpoint mixture at SCR's measured workload.
Controller time is cumulative per eight-environment batch.}
\label{fig:overview}\label{fig:tradeoff}
\end{figure*}

A visual world model lets a controller compare possible futures
before it acts. In image-goal planning, DINO-WM and LeWorldModel
(LeWM) score an action sequence by comparing its predicted terminal
representation with the encoded goal image~\cite{dinowm,lewm}.
Consider a robot pushing a block. Two plans can predict similar
final poses while proposing different initial pushes(Fig.~\ref{fig:overview}(a)). Another round
of the cross-entropy method (CEM)~\cite{cem} may resolve that
disagreement. It may also lower the predicted terminal cost while
barely changing the first push. In our controller, continuing from
five to fifteen iterations costs $10\times300=3{,}000$ additional
candidate evaluations over the same 25-command horizon.

Existing approaches spend this computation in different ways.
SufficientPlan selects a budget for a model--task pair using paired
closed-loop evaluations~\cite{sufficient}. LePlanner trains a
controller that performs a fixed number of learned plan updates
through a frozen predictor~\cite{leplanner}. Zhang et al.~\cite{zhang}
stop CEM when the smoothed mean elite cost shows little relative
improvement for several iterations. We study a single continuation
decision within each solve, using both progress in the cost of
complete plans and agreement about the action chunk that will be
executed first.

Selective CEM Refinement (SCR) makes that decision at iteration five
(Fig.~\ref{fig:overview}(b)). A logistic gate reads the displacement
of the first-chunk proposal mean from iterations four to five, the
relative decrease in mean elite terminal cost from iterations three
to five, and the dispersion of the fifth-iteration elites' first
chunks. It either releases the plan for execution or continues from
the current proposal through iteration fifteen. Both branches use
the same frozen encoder and predictor, sampling rule, population
size, horizon, and execution cap. On an early exit, SCR draws and
discards the random tensors for the skipped iterations. Their cost
is included in controller time, and no candidate from those draws
is evaluated by the world model.

The plan executed now can determine whether a later solve is needed.
We therefore count every scored candidate and time controller calls
over complete episodes. On OGBench-Cube~\cite{ogbench}, a fixed-depth
comparison yielded 25/32 successes at both five and thirty
iterations, while mean batch time rose from 2.196 to 7.313 seconds.
We evaluate the learned gate on PushT~\cite{pusht}, pairing trials
by source scene, starting frame, and planning seed.
Figure~\ref{fig:tradeoff}(c) shows the five measured controllers
alongside the analytic mixture of the two fixed policies. To locate
the savings from gating at replanning, we compare two controllers
with the same random-draw padding and identical initial decisions.
The same 14 trials reached a second solve. Applying the gate again
stopped nine solves early and saved 27,000 candidate evaluations
($15.79\%$ of total workload relative to initial-only gating).
Success fell from 27/32 to 25/32.

\section{Related Work}

\subsection{Visual world models and latent-space planning}

Learned dynamics can guide action selection at test time or provide trajectories for policy learning. PETS propagates uncertainty from an ensemble of probabilistic dynamics models through sampled trajectories during model predictive control~\cite{pets}. TD-MPC and TD-MPC2 optimize short action sequences in latent space and use a learned terminal value to account for rewards beyond the rollout horizon~\cite{tdmpc,tdmpc2}. Dreamer and DreamerV3 train policies on trajectories imagined by learned world models~\cite{dreamer, hafner2025dreamerv3}.

Finn and Levine used action-conditioned video prediction to plan robot pushes~\cite{finn2017deep}. Ebert et al. used image registration to track objects across observations so that a robot could retry a manipulation~\cite{ebert2018robustness}. For image-goal control, DINO-WM learns action-conditioned dynamics over frozen visual features and searches for actions that bring the predicted representation close to the goal~\cite{dinowm}. LeWorldModel (LeWM) jointly trains a visual encoder and an action-conditioned predictor from pixels using a next-representation prediction loss and SIGReg regularization~\cite{lewm,lejepa}. Its roughly 15-million-parameter model scores each plan by the distance between its predicted endpoint and the encoded goal. Maes et al.~\cite{lewm} also compare planning performance on PushT and OGBench-Cube at matched FLOPs. We hold the task-specific checkpoint and the prediction horizon fixed and ask a narrower question: after five iterations on the current observation, when is it worth spending ten more iterations refining the plan?

\subsection{Efficient sampling-based planning}

The cross-entropy method (CEM) repeatedly samples candidate plans and fits its proposal to the lowest-cost candidates~\cite{rubinstein1999cross, cem}. iCEM improves sample efficiency through temporally correlated sampling noise and elite reuse~\cite{icem}. CEM-GD applies gradient descent to promising sampled trajectories~\cite{cemgd}, while model predictive path integral control weights sampled control perturbations by their trajectory costs~\cite{mppi}.

SufficientPlan uses paired closed-loop outcomes to select a search budget for a model--task pair. Its Static-Context Reuse component caches observation and goal representations across search iterations~\cite{sufficient}. Selective CEM Refinement (SCR) makes a decision at a different level. It retains the solver's candidate-generation procedure, population size, rollout horizon, and execution cap, and chooses the iteration depth separately for each planning event.

\subsection{Adaptive search and early termination}

Zhang et al.~\cite{zhang} study early termination of CEM on LeWM, including PushT and OGBench-Cube. Their main rule smooths the mean elite cost across iterations and stops when its relative improvement remains below a threshold for several consecutive iterations. They also evaluate stopping criteria based on changes in the proposal mean and variance. SCR makes one decision after iteration five. It combines improvement in the cost of complete plans with two measurements of the first action chunk, the part of the plan that will be executed first.

Lu and Maciejowski compute the time of the next control update~\cite{lu2020self}. Zhang et al. use an event trigger to decide when a visual predictive controller should solve its optimization problem again~\cite{zhang2025event}. Magnino et al. learn a budget policy that couples planning time to the duration of open-loop execution~\cite{magnino}. Labiosa and Hanna use uncertainty in a reactive policy to decide when to invoke a planner~\cite{labiosa2026whentoplan}. On PushT, Zeer et al. invoke beam search when simulating the nominal action chunk predicts a sufficient drop in target coverage~\cite{zeer}. SCR makes its decision inside an ongoing CEM solve, after five iterations and with a fixed execution cap.

\section{Method}


\subsection{Closed-loop planning with a frozen world model}

Let $t$ count executed environment commands, $r$ index planning events within an episode, and $j$ index CEM iterations within an event. A command has dimension $d_a$. Each action chunk contains $b$ commands and has dimension $D=bd_a$. A candidate plan $\mathbf A=(\mathbf A_0,\ldots,\mathbf A_{H-1})$ contains $H$ chunks. The controller executes up to the first $K$ chunks, with $1\le K\le H$. In PushT, $d_a=2$ and $D=10$. The general definition of $D$ uses the action dimension of the task being evaluated.

At event $r$, the controller reads the current image $o_t$ and the goal image $o_g$. Let $E_\phi$ be the frozen encoder and $f_\theta$ the frozen action-conditioned predictor. Writing $\operatorname{Roll}_{f_\theta}$ for the checkpoint's rollout with its observation context, the cost of a complete plan is~\cite{lewm}
\begin{align}
 z_t &= E_\phi(o_t), \qquad z_g=E_\phi(o_g), \nonumber\\
 \hat z_H(\mathbf A) &= \operatorname{Roll}_{f_\theta}(z_t,\mathbf A), \nonumber\\
 \ell(\mathbf A) &= \|\hat z_H(\mathbf A)-z_g\|_2^2 .
 \label{eq:cost}
\end{align}
Thus each candidate is scored by the squared Euclidean distance between its predicted terminal representation and the goal representation.

The main protocol sets $b=H=K=5$ and allows $T=50$ commands per episode. Both a 5-iteration solve and a 15-iteration solve therefore look ahead 25 commands. Search leaves the environment at its current state. Once execution starts, the simulator checks success after every command and ends the episode as soon as the goal is reached. All episodes plan at $t=0$. Only episodes still active at $t=25$ receive a new observation and plan a second time. This is the feedback path in Fig.~\ref{fig:overview}(b).

\subsection{What the fifth iteration measures}

At iteration $j$, a Gaussian draw before native candidate processing is
\begin{align}
 \mathbf U_i^{(j)}
 &=\boldsymbol\mu^{(j-1)}
   +\boldsymbol\sigma^{(j-1)}\odot\boldsymbol\epsilon_i^{(j)}, \nonumber\\
 \boldsymbol\epsilon_i^{(j)}&\sim\mathcal N(0,I_{HD}).
 \label{eq:proposal}
\end{align}
Here $\boldsymbol\mu$ and $\boldsymbol\sigma$ are the coordinatewise proposal means and standard deviations, $\odot$ denotes coordinatewise multiplication, and $I_{HD}$ is the identity in the flattened plan space. The solver submits $M=300$ complete candidates for scoring by Eq.~\eqref{eq:cost}. Let $\mathcal E_j$ index the $L=30$ lowest-cost candidates, and let $\boldsymbol\mu^{(j)}$ denote the proposal mean after the elite update. The mean elite cost is
\begin{equation}
 \bar c_j=\frac{1}{L}\sum_{i\in\mathcal E_j}\ell(\mathbf A_i^{(j)}).
 \label{eq:elite}
\end{equation}

The subscript $0$ on $\mathbf A_{i,0}^{(j)}$ denotes the first chunk within a candidate plan at \emph{any} planning event. For its coordinate $k$, let
$\bar a_k=L^{-1}\sum_{i\in\mathcal E_5} A_{i,0,k}^{(5)}$.
The across-elite sample variance is
\begin{equation}
 V_k=\frac{1}{L-1}\sum_{i\in\mathcal E_5}
           (A_{i,0,k}^{(5)}-\bar a_k)^2 .
 \label{eq:var}
\end{equation}

Table~\ref{tab:notation} defines the three gate inputs. The displacement $d_5$ measures how far the proposed first chunk moved between iterations four and five. The spread $s_5$ measures how far fifth-iteration elites disagree on that chunk. The improvement $g_5$ uses the terminal costs of \emph{complete} plans and is positive when those costs fall. A solver can therefore exhibit a declining terminal cost while its elites still disagree about what to execute first.

\begin{table}[t]
\centering\small
\caption{Planning settings and fifth-iteration gate inputs. The $10^{-6}$ term stabilizes the relative cost change near zero.}
\label{tab:notation}
\begin{tabularx}{\columnwidth}{@{}l>{\raggedright\arraybackslash}X@{}}
\toprule
\multicolumn{2}{@{}l}{\textit{Planning and execution}}\\
$b=5,\ H=5,\ K=5$ & Chunk size, horizon, execution cap\\
$T=50$ & Command limit per episode\\
$M=300,\ L=30$ & Candidates and retained elites per iteration\\
\midrule
\multicolumn{2}{@{}l}{\textit{Statistics at the decision point}}\\
$d_5=\|\boldsymbol\mu^{(5)}_0-\boldsymbol\mu^{(4)}_0\|_2/\sqrt D$
  & First-chunk mean displacement\\
$g_5=(\bar c_3-\bar c_5)/(|\bar c_3|+10^{-6})$
  & Full-plan elite cost improvement\\
$s_5=\sqrt{D^{-1}\sum_{k=1}^D V_k}$
  & First-chunk elite spread\\
\bottomrule
\end{tabularx}
\end{table}


\subsection{One continuation decision per solve}

Let $\mathbf x=(d_5,g_5,s_5)^\top$. For feature $k$, $m_k$ and $v_k$ are its development-set mean and standard deviation. The logistic gate has weights $\mathbf w$ and intercept $\beta_0$. A higher score favors continuation:
\begin{align}
 \tilde x_k &= (x_k-m_k)/v_k,\qquad k=1,2,3, \nonumber\\
 p(\mathbf x)&=
   \bigl(1+\exp[-(\beta_0+\mathbf w^\top\tilde{\mathbf x})]\bigr)^{-1},
 \label{eq:score}\\
 J(\mathbf x)&=
 \begin{cases}
 15,&p(\mathbf x)\ge\tau,\\
 5,&p(\mathbf x)<\tau ,
 \end{cases}
 \qquad \tau\simeq 0.3966121994 .
 \label{eq:decision}
\end{align}

A solve that continues retains its proposal and runs iterations 6 through 15. There is no further gate evaluation in that solve. Both branches hand the selected plan to the same execution routine. In particular, an early exit changes how much computation precedes execution, while the command horizon and the execution cap remain the same.

The score in Eq.~\eqref{eq:score} ranks continuation decisions. Interpreting the benefit of continuation at $t=25$ requires comparing the two branches from the same observed state and five-iteration search history. The fixed-depth controllers can reach different states at that time. Their paired whole-episode outcomes therefore cannot supply counterfactual training labels for an individual replanning event.


\subsection{Random draws and closed-loop workload}

Yadav et al. use common random numbers to compare rollout estimates of action values~\cite{yadav2026common}. In our padded variants, a solve that stops at iteration five draws and discards the random tensors for the ten skipped iterations. Controller time includes these draws. The candidate count $W_m$ includes only plans scored by the world model. Padding matches the nominal draw count within a solve. When episodes terminate at different times, the order of active environments can still change which draws reach later solves.

For method $m$, let $N_{m,5}$ and $N_{m,15}$ count the solves that actually end at each depth. Total solves $N_m$, CEM iterations $R_m$, and evaluated candidates $W_m$ satisfy
\begin{align}
 N_m &= N_{m,5}+N_{m,15}, \nonumber\\
 R_m &= 5N_{m,5}+15N_{m,15}, \qquad W_m=MR_m .
 \label{eq:work}
\end{align}
A shallow solve evaluates 1,500 complete plans and a deep solve evaluates 4,500. Each batch contains eight environments. The solver visits active environments sequentially and scores the 300 candidates for one environment together. Equation~\eqref{eq:work} counts evaluated plans across all actual planning events, including any additional replanning induced by a controller's earlier actions.

\section{Experiments}
\label{sec:experiments}

\subsection{Environments and evaluation protocol}

We study test-time planning with frozen, task-specific LeWorldModel
(LeWM) checkpoints~\cite{lewm} on planar PushT~\cite{pusht} and the
single-cube manipulation task from OGBench~\cite{ogbench}. We use the PushT and single-cube environment implementations provided by \texttt{stable-worldmodel}~\cite{stablewm}. On both
tasks, CEM~\cite{cem} ranks each action sequence by the squared distance
between its predicted terminal representation and the encoded goal image.
The fixed-depth experiments measure the return from additional search
on each task. The five-controller comparison, stage ablation, and
paired outcome analysis evaluate SCR on PushT. Transfer of the final
SCR gate to Cube remains untested.

\begin{table*}[t]
\centering\small
\caption{Task configurations. Development seeds select the scenes used
for the fixed-depth sweep. The five-controller comparison uses a
separate PushT cohort.}
\label{tab:exp_setup}
\begin{tabular}{lll}
\toprule
Setting & PushT & OGBench-Cube\\
\midrule
Environment
  & \texttt{swm/PushT-v1}
  & \texttt{swm/OGBCube-v0} (single cube)\\
Dataset
  & \texttt{pusht\_expert\_train}
  & \texttt{ogbench/cube\_single\_expert}\\
Frozen checkpoint
  & \texttt{native\_pusht/weights.pt}
  & \texttt{native\_cube/weights.pt}\\
Use in this study
  & Fixed depths and SCR & Fixed depths only\\
Development depths $J$ & 5, 15, 30 & 5, 30\\
Development scene seeds & 72011, 72012 & 72011, 72012\\
Main scene seeds & 93011, 93012 & ---\\
CEM seeds & 42, 99 & 42, 99\\
\bottomrule
\end{tabular}
\end{table*}

Each action chunk contains $b=5$ commands. A candidate spans $H=5$
chunks, and the controller executes at most $K=5$ chunks before
replanning. CEM scores $M=300$ complete candidates and retains $L=30$
elites per iteration. Episodes end on success or after $T=50$ commands.
The dataset goal offset and the execution interval both equal 25
commands. The offset selects the target frame in the recorded episode.
The interval $Kb$ determines when an active environment receives a
new plan.

For each task and protocol in Table~\ref{tab:exp_setup}, two scene
seeds each select eight source scenes. Each scene is evaluated with
CEM seeds 42 and 99, yielding 32 trials from 16 distinct scenes per
controller. The four main PushT batches, denoted A--D, use
$(93011,42)$, $(93011,99)$, $(93012,42)$, and $(93012,99)$,
respectively. Each pair specifies the scene seed followed by the CEM
seed. Controllers are paired by source episode, starting frame, and
CEM seed. The main scenes are disjoint from those used to calibrate the
gate. The padding variants were subsequently evaluated on this same
cohort during mechanism analysis, which limits the independence of
that comparison.

\subsection{Evaluation metrics and timing}

Let $Y_{m,i}\in\{0,1\}$ denote success for controller $m$ on trial $i$.
The simulator checks the task criterion after each executed command.
We report the empirical success rate
\begin{equation}
  \widehat S_m=\frac{1}{n}\sum_{i=1}^{n}Y_{m,i},
  \qquad n=32 .
  \label{eq:exp_success}
\end{equation}

For the five-controller comparison, let $N_{m,5}$ and $N_{m,15}$ count
the planning events that end after five and fifteen CEM iterations,
respectively. With $N_m=N_{m,5}+N_{m,15}$, the closed-loop counts in
Eq.~\eqref{eq:work} give
\begin{equation}
  R_m=5N_{m,5}+15N_{m,15},\qquad W_m=300R_m .
  \label{eq:exp_work}
\end{equation}
The unit of $W_m$ is one complete candidate plan scored by the world
model. Discarded padding draws add elapsed time and contribute zero
candidate evaluations.

Let $u_{m,h}$ be \texttt{control\_wall\_s}, the time accumulated over
all controller calls in eight-environment batch $h$. We report
$\bar u_m\pm s_m$ in seconds per batch, where
\begin{equation}
  \bar u_m=\frac14\sum_{h=1}^{4}u_{m,h},\qquad
  s_m=\sqrt{\frac{1}{3}\sum_{h=1}^{4}(u_{m,h}-\bar u_m)^2}.
  \label{eq:exp_time}
\end{equation}
Timing used an NVIDIA RTX PRO 4000 Blackwell GPU with
PyTorch~2.8.0+cu128. We synchronized CUDA before and after each timed
controller call. The reported spread is the sample standard deviation
of four batch totals. Per-action latency was not measured.

\subsection{Returns from fixed search depth}

Table~\ref{tab:fixed_depth} shows where deeper search helped. On the
PushT development cohort, increasing $J$ from 5 to 15 raised success
from 15/32 to 30/32. Increasing $J$ from 15 to 30 took a further
2.820 seconds per batch and yielded 29/32 successes.

On Cube, both depths succeeded in 25/32 trials, while mean batch time
rose from 2.196 to 7.313 seconds. Five iterations used about 30\% of
the controller time required by 30 iterations. The two controllers
succeeded together on 22 trials. Each succeeded on three trials where
the other failed, and both failed on the remaining four. Search depth
therefore changed individual Cube outcomes even though the aggregate
success count was unchanged. The PushT development results and the
main results below use different source scenes.

\begin{table}[t]
\centering\small
\setlength{\tabcolsep}{3.5pt}
\caption{Fixed depths on development scenes. Each row contains 32
trials. Time is accumulated controller time per batch, reported as
mean $\pm$ sample standard deviation.}
\label{tab:fixed_depth}
\begin{tabular}{lrrrr}
\toprule
Task & $J$ & Successes & Success (\%) & Time (s)\\
\midrule
PushT & 5  & 15/32 & 46.88 & $2.497\pm0.090$\\
PushT & 15 & 30/32 & 93.75 & $3.775\pm0.163$\\
PushT & 30 & 29/32 & 90.63 & $6.595\pm0.350$\\
\midrule
Cube  & 5  & 25/32 & 78.13 & $2.196\pm0.085$\\
Cube  & 30 & 25/32 & 78.13 & $7.313\pm0.619$\\
\bottomrule
\end{tabular}
\end{table}

\subsection{Main PushT results}

Fixed-5 and Fixed-15 use the same depth at every solve. The initial
gate (IG) makes one decision after iteration five at $t=0$ and always
uses 15 iterations at replanning. IG+P adds random-draw padding after
an early exit. SCR applies the same padded gate at every planning
event. Padding generates and discards the random tensors for the
skipped iterations without scoring additional trajectories.

\begin{table*}[t]
\centering\small
\caption{Main PushT results on 32 paired trials. $N_5/N_{15}$ counts
solves at each stopping depth. $R$ and $W$ are totals over all 32
trials. Time is the mean and sample standard deviation of four batch
totals.}
\label{tab:main}
\begin{tabular}{lrrrrrrr}
\toprule
Controller & Successes & Success (\%) & Solves $N$ & $N_5/N_{15}$ &
CEM iterations $R$ & Candidates $W$ & Time (s/batch)\\
\midrule
Fixed-5  & 17/32 & 53.13 & 56 & 56/0  & 280 & 84,000  & $2.567\pm0.128$\\
Fixed-15 & 28/32 & 87.50 & 42 & 0/42  & 630 & 189,000 & $4.286\pm0.689$\\
IG       & 26/32 & 81.25 & 46 & 13/33 & 560 & 168,000 & $4.043\pm0.567$\\
IG+P     & 27/32 & 84.38 & 46 & 12/34 & 570 & 171,000 & $4.010\pm0.450$\\
SCR      & 25/32 & 78.13 & 46 & 21/25 & 480 & 144,000 & $3.548\pm0.286$\\
\bottomrule
\end{tabular}
\end{table*}

Relative to Fixed-15, SCR evaluated 45,000 fewer candidates
($23.81\%$) and saved 0.738 seconds per batch ($17.22\%$). Success
decreased by three trials, or 9.375 percentage points (pp). Relative
to Fixed-5, SCR gained eight successes ($+25.0$ pp) at a cost of
60,000 additional candidate evaluations ($+71.43\%$) and 0.981
additional seconds per batch ($+38.23\%$). Among the adaptive
controllers, IG+P achieved the most successes (27/32) and evaluated
171,000 candidates (Table~\ref{tab:main}).

\paragraph{Cost-matched endpoint mixture.}
The filled markers in Fig.~\ref{fig:tradeoff}(c) show the five
measured controllers. To interpret the dashed line, consider assigning
each complete eight-environment batch to Fixed-15 with probability $q$
and to Fixed-5 otherwise. For a cost measure $C$, let $C_5$ and
$C_{15}$ denote the two measured endpoint costs, and let $C_m$ denote
the cost of controller $m$. The resulting reference matches cost in
expectation:
\begin{align}
  q_C &=
  \frac{C_m-C_5}{C_{15}-C_5}, \nonumber\\
  \widehat S_{\mathrm{mix}}(q_C)
    &= (1-q_C)\widehat S_5+q_C\widehat S_{15}, \nonumber\\
  \Delta_C &=
  100\bigl[\widehat S_m-\widehat S_{\mathrm{mix}}(q_C)\bigr]
  \quad\text{(pp)} .
  \label{eq:exp_mix}
\end{align}

For SCR, matching $W=144{,}000$ gives $q_W=4/7$. The open circle in
Fig.~\ref{fig:tradeoff}(c) marks the implied $72.768\%$ success rate.
SCR achieved $25/32=78.125\%$, a difference of $+5.357$ pp. The
workload-matched references for IG and IG+P are $80.625\%$ and
$81.607\%$ ($q_W=0.8000$ and $0.8286$), giving differences of
$+0.625$ and $+2.768$ pp. These references interpolate the two
measured endpoints. Intermediate fixed depths and cost-matched
random gates were not evaluated on this cohort.

\subsection{Stage ablation}

All 32 trials require an initial solve. At $t=25$, Fixed-5 replans
24 times, Fixed-15 replans 10 times, and each adaptive controller
replans 14 times (Table~\ref{tab:stages}). Solve counts and success
counts use different units. For example, SCR batch A made eight
initial and five subsequent solves, while six of its eight trials
succeeded.

\begin{table*}[t]
\centering\small
\caption{Stage-resolved PushT workload. Each 5/15 entry gives the
number of five-iteration and fifteen-iteration solves. $W_0$ and
$W_{25}$ count evaluated candidates at the two stages. The final two
columns count successful trials.}
\label{tab:stages}
\begin{tabular}{lccrrrrr}
\toprule
Controller & $t=0$ (5/15) & $t=25$ (5/15) &
$W_0$ & $W_{25}$ & Total $W$ & By $t=25$ & Final\\
\midrule
Fixed-5  & 32/0  & 24/0 & 48,000  & 36,000 & 84,000  & 8  & 17\\
Fixed-15 & 0/32  & 0/10 & 144,000 & 45,000 & 189,000 & 22 & 28\\
IG       & 13/19 & 0/14 & 105,000 & 63,000 & 168,000 & 18 & 26\\
IG+P     & 12/20 & 0/14 & 108,000 & 63,000 & 171,000 & 18 & 27\\
SCR      & 12/20 & 9/5  & 108,000 & 36,000 & 144,000 & 18 & 25\\
\bottomrule
\end{tabular}
\end{table*}

\paragraph{Padding at the initial solve.}
IG and IG+P both applied the gate only at $t=0$. Adding random-draw
padding raised success from $26/32$ to $27/32$ and retention of
Fixed-15 successes from $25/28$ to $27/28$. Both variants solved
11 of the 12 Fixed-15-only trials. Among the 16 trials solved by
both fixed controllers, IG solved 14 and IG+P solved all 16.
IG also solved the single Fixed-5-only trial, which IG+P missed.
Total candidate evaluations rose from 168,000 to 171,000.
Table~\ref{tab:replan_pairs} summarizes the three adaptive controllers.

\paragraph{Gating at replanning.}
IG+P and SCR had identical recorded features, scores, and decisions
at the initial solve. Both evaluated 108,000 candidates at $t=0$,
succeeded on the same 18 trials before replanning, and replanned on
the remaining 14 at $t=25$. IG+P ran all 14 replans to iteration 15.
SCR stopped nine at iteration five and continued the other five.
These nine exits saved $9(15-5)\times300=27{,}000$ candidate
evaluations. Replanning work fell from 63,000 to 36,000
($42.86\%$), and total work fell from 171,000 to 144,000
($15.79\%$). Mean controller time per eight-environment batch fell
from 4.010 to 3.548 seconds, a saving of 0.462 seconds ($11.53\%$).
Of the 14 replanned trials, IG+P solved nine and SCR solved seven.
All seven SCR successes were shared. Over all 32 trials, success
therefore fell from $27/32$ to $25/32$ ($-6.25$ percentage points).

\begin{table}[t]
\centering\small
\setlength{\tabcolsep}{2.6pt}
\caption{PushT stage ablation. In panel (a), the success columns count
all successes and those shared with Fixed-15. Savings are relative
to Fixed-15. Each gap is the observed success rate minus the endpoint
mixture matched to that controller's candidate workload ($W$) or
mean batch controller time. $R/N$ is the mean number of CEM iterations
per solve. Panel (b) counts final paired outcomes.}
\label{tab:replan_pairs}
\begin{tabular}{@{}lrrrrrrr@{}}
\toprule
\multicolumn{8}{@{}l}{\textit{(a) All 32 trials}}\\
 & \multicolumn{2}{c}{Successes} & &
   \multicolumn{2}{c}{Saved (\%)} & \multicolumn{2}{c}{Gap (pp)}\\
\cmidrule(lr){2-3}\cmidrule(lr){5-6}\cmidrule(l){7-8}
Gate & /32 & /28 & $R/N$ & $W$ & Time & $W$ & Time\\
\midrule
IG   & 26 & 25 & 12.17 & 11.11 &  5.67 & $+0.625$ & $-1.394$\\
IG+P & 27 & 27 & 12.39 &  9.52 &  6.43 & $+2.768$ & $+2.386$\\
SCR  & 25 & 25 & 10.43 & 23.81 & 17.22 & $+5.357$ & $+5.379$\\
\bottomrule
\end{tabular}
\par\smallskip
\begin{tabular}{@{}lrrr@{}}
\toprule
\multicolumn{4}{@{}l}{\textit{(b) The same 14 trials reached replanning}}\\
 & \multicolumn{2}{c}{SCR} & \\
\cmidrule(lr){2-3}
IG+P & Success & Failure & Total\\
\midrule
Success & 7 & 2 & 9\\
Failure & 0 & 5 & 5\\
\midrule
Total   & 7 & 7 & 14\\
\bottomrule
\end{tabular}
\end{table}

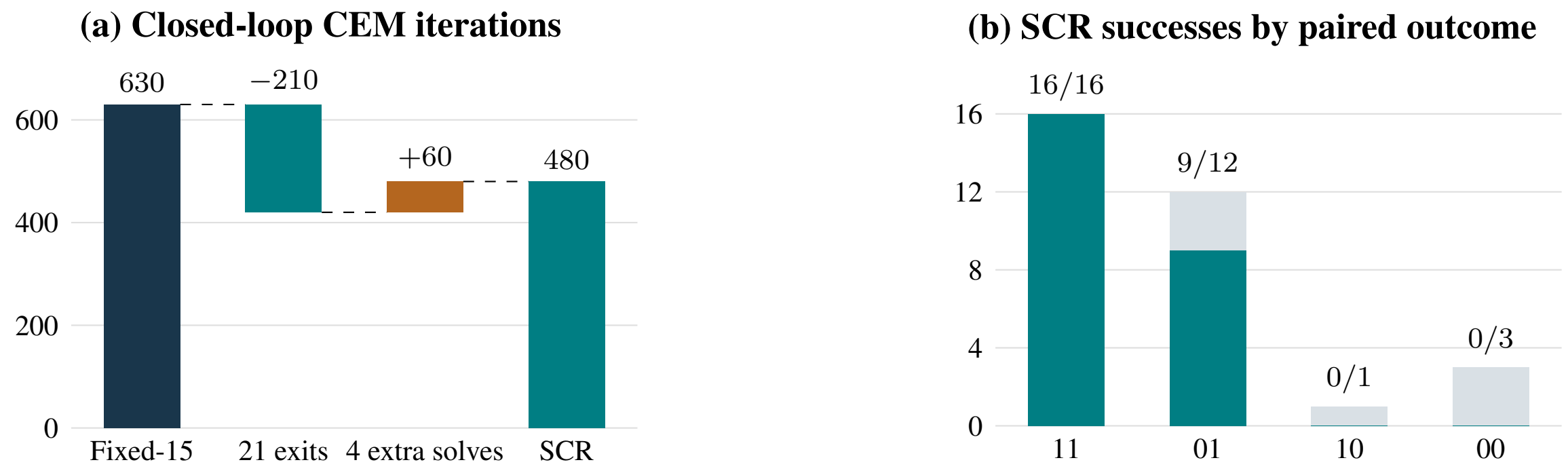
\begin{figure*}[t]
\centering
\sourcefigure{166FA91C-80FF-4109-8984-F7788B017FB3}{%
\begin{minipage}[c]{.48\textwidth}
\centering
\textbf{(a) Closed-loop CEM iterations}\par\smallskip
\begin{tikzpicture}[x=1.38cm,y=.005cm,font=\footnotesize]
  \foreach \y in {0,200,400,600}{
    \draw[gray!25] (.5,\y)--(4.5,\y);
    \node[left] at (.5,\y) {\y};
  }
  \foreach \x/\lo/\hi/\col/\val in
    {1/0/630/ink/630,2/420/630/scrteal/-210,
     3/420/480/warm/+60,4/0/480/scrteal/480}{
    \fill[\col] (\x-.27,\lo) rectangle (\x+.27,\hi);
    \node[above] at (\x,\hi) {$\val$};
  }
  \draw[dashed] (1.27,630)--(1.73,630)
                (2.27,420)--(2.73,420)
                (3.27,480)--(3.73,480);
  \foreach \x/\lab in
    {1/Fixed-15,2/21 exits,3/4 extra solves,4/SCR}
    \node[below] at (\x,0) {\lab};
\end{tikzpicture}
\end{minipage}\hfill
\begin{minipage}[c]{.48\textwidth}
\centering
\textbf{(b) SCR successes by paired outcome}\par\smallskip
\begin{tikzpicture}[x=1.38cm,y=.19cm,font=\footnotesize]
  \foreach \y in {0,4,8,12,16}{
    \draw[gray!25] (.5,\y)--(4.5,\y);
    \node[left] at (.5,\y) {\y};
  }
  \foreach \x/\n/\s/\lab in
    {1/16/16/11,2/12/9/01,3/1/0/10,4/3/0/00}{
    \fill[pale] (\x-.27,0) rectangle (\x+.27,\n);
    \fill[scrteal] (\x-.27,0) rectangle (\x+.27,\s);
    \node[above] at (\x,\n) {$\s/\n$};
    \node[below] at (\x,0) {\lab};
  }
\end{tikzpicture}
\end{minipage}}
\caption{(a) Early exits save 210 CEM iterations. The four additional
solves add back 60 iterations at the reference depth, leaving 480
iterations for SCR. (b) Groups are defined by the paired outcomes of
Fixed-5 and Fixed-15, in that order. Each fraction gives SCR successes
over the number of trials in the group.}
\label{fig:audit}
\end{figure*}

Figure~\ref{fig:audit}(a) reconciles SCR's 21 early exits with its
net savings against Fixed-15. Counting each additional solve at the
15-iteration reference depth gives
\begin{equation}
  R_{\mathrm{SCR}}
  =630-\underbrace{21(15-5)}_{210}
      +\underbrace{(46-42)15}_{60}
  =480 .
  \label{eq:exp_reconcile}
\end{equation}
The net saving is 150 iterations, or 45,000 candidate evaluations.
The $+60$ term accounts for the four additional solves under SCR.
The two initial-only gates also differed across batches. IG and IG+P
both made 46 solves, distributed as $(12,14,11,9)$ and $(13,13,11,9)$,
respectively, across batches A--D.

\subsection{Paired success retention}

Table~\ref{tab:paired} groups trials by the observed pair
$(Y_5,Y_{15})$. SCR succeeded in all 16 trials where both fixed
controllers succeeded and in nine of the 12 trials where only Fixed-15
succeeded. Its three misses in the latter group were A4, B4, and B7.
The single Fixed-5-only trial and the three mutual failures also
failed under SCR (Fig.~\ref{fig:audit}(b)).

\begin{table}[t]
\centering\small
\setlength{\tabcolsep}{3.3pt}
\caption{Paired PushT outcomes. Controller columns give successes
within each group. F5 and F15 denote the fixed controllers. The last
row counts successes shared with F15.}
\label{tab:paired}
\begin{tabular}{crrrrrr}
\toprule
$(Y_5,Y_{15})$ & Trials & F5 & F15 & IG & IG+P & SCR\\
\midrule
$(1,1)$ & 16 & 16 & 16 & 14 & 16 & 16\\
$(0,1)$ & 12 & 0  & 12 & 11 & 11 & 9\\
$(1,0)$ & 1  & 1  & 0  & 1  & 0  & 0\\
$(0,0)$ & 3  & 0  & 0  & 0  & 0  & 0\\
\midrule
Total & 32 & 17 & 28 & 26 & 27 & 25\\
\multicolumn{2}{l}{Retained of 28} & 16 & 28 & 25 & 27 & 25\\
\bottomrule
\end{tabular}
\end{table}

We define retention as the fraction of Fixed-15 successes that the
other controller also achieves:
\begin{equation}
  \operatorname{Ret}_{m\mid15}
  =\frac{\sum_{i=1}^{n}Y_{m,i}Y_{15,i}}
         {\sum_{i=1}^{n}Y_{15,i}} .
  \label{eq:exp_retention}
\end{equation}
SCR retained $25/28=89.29\%$ of the observed Fixed-15 successes.
Its absolute task success rate was $25/32=78.125\%$. IG succeeded on
26 trials, including A3, where Fixed-15 failed. Its retention
numerator is therefore 25.

The paired groups also explain the workload-matched contrast. The
mutual-success and mutual-failure groups contribute zero to the
difference. At $q_W=4/7$, the Fixed-15-only group contributes
$9-12q_W$ successes above the mixture expectation. The Fixed-5-only
group contributes $0-(1-q_W)$. Together these terms yield $12/7$
expected success equivalents across 32 trials, or $+5.357$ pp.
Group membership depends on the planning seed. For example, A1 and
B1 use the same source scene and belong to $(0,1)$ and $(0,0)$,
respectively.


\subsection{Planning seeds and statistical uncertainty}

Table~\ref{tab:exp_seeds} recomputes the cost-matching weight within
each planning seed. The workload-matched contrast was $+11.486$ pp
for seed 42 and $-1.136$ pp for seed 99. Using mean batch controller
time as $C$ in Eq.~\eqref{eq:exp_mix} gave $+12.289$ and $-1.617$ pp,
respectively.

\begin{table}[t]
\centering\small
\setlength{\tabcolsep}{3.3pt}
\caption{Seed-stratified PushT results. Success columns give counts
out of $n$. The final two columns give SCR minus the endpoint mixture
matched on workload or controller time, in percentage points.}
\label{tab:exp_seeds}
\begin{tabular}{lrrrrrr}
\toprule
Seed & $n$ & F5 & F15 & SCR & $\Delta_W$ & $\Delta_{\mathrm{time}}$\\
\midrule
42     & 16 & 8  & 15 & 14 & $+11.486$ & $+12.289$\\
99     & 16 & 9  & 13 & 11 & $-1.136$  & $-1.617$\\
Pooled & 32 & 17 & 28 & 25 & $+5.357$  & $+5.379$\\
\bottomrule
\end{tabular}
\end{table}

At SCR's workload of 144,000 candidate evaluations, the endpoint
mixture assigned weight $q_W=4/7$ to Fixed-15. Its expected success
rate was $72.768\%$, compared with $25/32=78.125\%$ for SCR, a
difference of $+5.357$ percentage points (pp). Averaging the two
planning seeds within each source scene, SCR exceeded the mixture on
six scenes, tied it on seven, and fell below it on three. Leaving out
one scene at a time gave differences from $+2.86$ to $+9.52$ pp.
An exact percentile bootstrap that resampled 16 source scenes with
replacement, keeping both planning seeds together, gave a conditional
95\% interval of $[-6.70,+16.07]$ pp. Both analyses held $q_W$ fixed.

Matching mean controller time per batch gave $q_{\rm time}=0.570796$
and a pooled difference of $+5.379$ pp. For this comparison, 200,000
source-scene bootstrap replicates gave a conditional 95\% interval of
$[-6.679,+16.551]$ pp, with both seeds retained for each sampled
scene and the measured times and mixture weight held fixed. Shared
random draws can couple environments within a batch, a dependence
that scene resampling does not capture. Both intervals include zero,
and the observed difference changes sign across planning seeds under
both matching rules.

\FloatBarrier
\section{Conclusion}
\label{sec:conclusion}

Selective CEM Refinement (SCR) decides whether to execute the plan
available after five CEM iterations or to refine it for ten more.
The gate uses the change in the proposed first action chunk, the
improvement in the terminal cost of elite plans, and the spread of
first chunks among elites. All three statistics come from candidates
already scored by the frozen world model, so the decision requires
no additional model rollout.

The return from deeper search differed across tasks. On the PushT
development scenes, increasing the depth from 5 to 15 raised success
from 15/32 to 30/32. On Cube, depths of 5 and 30 each solved 25/32
trials, while mean controller time per batch rose from 2.196 to
7.313 seconds (Table~\ref{tab:fixed_depth}).

In the main PushT comparison, five controllers ran on the same 32
scene--seed pairs, for 160 runs and 756,000 candidate evaluations
in total. SCR solved 25/32 trials ($78.13\%$). All 25 were among
the 28 solved by Fixed-15, giving a paired retention rate of
$89.29\%$ (Table~\ref{tab:main}). SCR stopped after five iterations
on nine of its 14 replans. Even with four more solves than Fixed-15
over the complete episodes, SCR reduced candidate evaluations from
189,000 to 144,000 ($23.81\%$) and mean controller time per
eight-environment batch from 4.286 to 3.548 seconds, a saving of
0.738 seconds ($17.22\%$). These totals include the cost of every
replan (Table~\ref{tab:stages}). At SCR's workload, the analytic
mixture of Fixed-5 and Fixed-15 implies an expected success rate of
$72.77\%$. SCR's observed rate was 5.36 percentage points higher
(Fig.~\ref{fig:tradeoff}(c)).

\section{Limitations and future work}
\label{sec:limitations}

\paragraph{Evidence at matched cost.}
The main evaluation covers 16 source scenes and two planning seeds.
The workload-matched difference was $+11.49$ percentage points for
seed 42 and $-1.14$ for seed 99. Matching mean controller time gave
a pooled difference of $+5.38$ points, with a conditional 95\%
source-cluster bootstrap interval of $[-6.68,+16.55]$ points
(Table~\ref{tab:exp_seeds}). This interval applies to the
time-matched comparison and holds the measured times and mixture
weight fixed. Whether the gate improves success at equal cost on
new scenes remains unresolved. A direct test would compare SCR on
new scenes and planning seeds with an intermediate fixed depth,
a feature-independent random gate matched in expected workload,
and an elite-cost stopping rule~\cite{zhang}, all under the same
execution protocol.

\paragraph{Supervision at replanning.}
Gate fitting uses whole-episode outcomes as a proxy for the value of continuing the initial solve. By the second solve, the two fixed-depth controllers may already have reached different states. An event-level comparison could branch from the same simulator state, CEM proposal, and random generator state immediately after iteration five. One branch would execute the current plan, and the other would refine it through iteration 15. Both would then follow the same subsequent control policy. This would measure the effect of continuation at the state where the gate actually makes its decision. Random-draw padding accounts for skipped draws within one solve. A counter-based generator~\cite{salmon2011parallel} would let us key draws by trial and planning event and index them by CEM iteration, candidate, and action coordinate. This would keep one environment's termination from shifting the draws assigned to another.

\paragraph{Transfer and deployment.}
The learned gate was evaluated only on PushT. Vasilev and Nayak measured differences between LeWM and simulator responses to small action perturbations on Reacher and OGBench-Cube~\cite{vasilev2026does}. Whether the first-chunk spread $s_5$ predicts the benefit of further search on Cube remains an empirical question. A direct transfer test would apply the PushT feature normalization, gate coefficients, and threshold to Cube. A separate adaptation test would recalibrate the gate on Cube development scenes and evaluate it on held-out Cube scenes. The current timing metric aggregates controller calls over an eight-environment simulation batch. Per-command latency and energy use on a physical robot remain unmeasured.

\section*{LLM Usage Statement}
We developed the method, ran the experiments, selected the references, and wrote the original manuscript. We used OpenAI ChatGPT and Codex to
edit and translate passages from our drafts, check bibliographic details and reported calculations, and we also used OpenAI ChatGPT and Google Gemini for reviewer-style feedback.
We take responsibility for the final manuscript and its claims.
{\small
\bibliographystyle{ieeenat_fullname}
\bibliography{main}
}

\end{document}